\documentclass[runningheads]{llncs}

\usepackage{graphicx}
\usepackage{subcaption}
\usepackage{booktabs}
\usepackage{amsmath}

\usepackage{amsthm}
\usepackage{amssymb}

\usepackage{algorithm,algorithmicx,algpseudocode}
\usepackage{multirow}
\usepackage{float}
\usepackage{bm}
\usepackage{hyperref}
\usepackage[T1]{fontenc}

\usepackage{pgfplots}
\pgfplotsset{compat=1.9}

\begin{document}
\title{Influence Based Defense Against Data Poisoning Attacks in Online Learning}
%
%

\author{Sanjay Seetharaman \and Shubham Malaviya \and Rosni KV \and Manish Shukla \and Sachin Lodha}

\authorrunning{Seetharaman et al.}
%
\institute{TCS Research, India\\
\email{\{s.seetharaman1,shubham.malaviya,rosni.kv,mani.shukla,sachin.lodha\}@tcs.com}}

\maketitle
\begin{abstract}
    Data poisoning is a type of adversarial attack on training data where an attacker manipulates a fraction of data to degrade the performance of machine learning model. Therefore, applications that rely on external data-sources for training data are at a significantly higher risk. There are several known defensive mechanisms that can help in mitigating the threat from such attacks. For example, data sanitization is a popular defensive mechanism wherein the learner rejects those data points that are sufficiently far from the set of training instances. Prior work on data poisoning defense primarily focused on offline setting, wherein all the data is assumed to be available for analysis. Defensive measures for online learning, where data points arrive sequentially, have not garnered similar interest. 
    
    In this work, we propose a defense mechanism to minimize the degradation caused by the poisoned training data on a learner's model in an online setup. Our proposed method utilizes an influence function which is a classic technique in robust statistics. Further, we supplement it with the existing data sanitization methods for filtering out some of the poisoned data points. We study the effectiveness of our defense mechanism on multiple datasets and across multiple attack strategies against an online learner.

\end{abstract}
\section{Introduction}
    Machine Learning (ML) plays a crucial part in a wide variety of applications. These applications are usually data-intensive and use ML algorithms for building a mathematical model from the training data for decision making. The training data used for building the model could come from controlled and uncontrolled data sources. Any imperfection, intentional or unintentional, in the data could lead to model corruption, and thus producing unexpected results post deployment. The introduction of intentional imperfection in the training data is commonly known as \emph{data poisoning}, which is a type of adversarial attack \cite{xiao2015feature} that degrades the performance of the trained classifier. As the adversary could manipulate the training data to subvert the learning process of the model or manipulate the test data to evade the model prediction, therefore it becomes imperative to secure the ML model against such attacks. 

    Two of the major attacking strategies are \emph{evasion} and \emph{poisoning} attacks. In an evasion attack, the test data is manipulated to evade the classifier's decision boundary. In a recent work, Labaca et al \cite{labaca2019poster} demonstrated the evasion attack on malware classifiers, wherein they generated valid adversarial malware samples against convolutional neural networks using gradient descent technique. In contrast to this, in a data poisoning attack, the training data is manipulated to achieve the attacker's objective. For example, Nelson et al \cite{nelson2008exploiting} have shown that even a 1\% perturbation in training data makes the spam-detection system useless. 

    Biggio et al \cite{biggio2012poisoning} described data poisoning attack as a causative attack in which training data is injected with especially crafted malicious data-points. An attacker can mount this attack either by directly modifying the training dataset (insider threat) \cite{homoliak2019insight} or by submitting malicious samples to the ML model \cite{biggio2012poisoning}. In past, anti-virus vendors have been blamed for injecting poisoned samples into VirusTotal\footnote{ https://www.virustotal.com/} for degrading the performance of the competing products \cite{biggio2018wild}. The machine learning models which use training data from unconstrained and unmonitored source have high susceptibility to data poisoning attacks. The twitter bot, \emph{Microsoft Tay}, is one such illustrious example, where a crowd, aged between 18 to 24, corrupted it to post inflammatory and offensive tweets within 16 hours after its deployment \cite{price2016microsoft}. Similarly, Lam et al \cite{lam2004shilling} studied the shilling attack in recommender systems wherein the untrustworthy \emph{crowd} could influence the system in recommending low-quality items to an unsuspecting user.

    Data poisoning attack is a serious threat to a ML model as it could result in security and privacy issues, monetary losses and other serious implications \cite{biggio2012poisoning,mei2015using,zhao2017efficient,koh2018stronger,zhang2019online}. The impact of data poisoning attack could be seen on a variety of applications, such as spam filtering \cite{nelson2008exploiting}, malware detection \cite{labaca2019poster}, recommender system \cite{lam2004shilling} and sentiment analysis \cite{newell2014practicality}. For example, Lowd and Meek \cite{lowd2005good} have shown that it is possible to fool a spam classifier by carefully composing the message body of the email. Similarly, it has been shown that a \emph{malicious producer} could change the outcome of a recommender system by inserting fake product reviews, such that their product is often recommended to customers~\cite{chen2019data}. 

    The existing literature on data poisoning primarily focuses on offline setting, wherein the entire data is available for training of the classifier and for mounting data poisoning attack \cite{biggio2012poisoning,mei2015using,zhao2017efficient}. However, these studies do not address the adversarial threat against an online learner. For example, consider an e-commerce application where the user generated data arrives sequentially. As the user is in control of data generation, it is relatively easy for a malicious user to manipulate the data sample before it reaches to the learner. Thus, making the online learner an easy target for data poisoning attack. Moreover, the adversary has the advantage of observing the impact of previous data-item she poisoned and accordingly adjusting her attack strategy~\cite{zhang2019online}.

    One of the well-known defense strategies against data poisoning attack is \emph{data sanitization} \cite{cretu2008casting}, where the suspicious data is filtered out before it reaches to the training process \cite{paudice2018label,chan2018data}. Steinhardt et al \cite{steinhardt2017certified} described an approximate upper bound on the maximum test error a defender (who sanitizes the data before training) can suffer under attack. In their work, Koh et al \cite{koh2018stronger} explored the data poisoning attacks against the common data sanitization defenses. They concluded that data sanitization defense tends to fail easily and discussed different strategies for stronger defenses. The other defense strategies against data poisoning prescribe training of model from reliable dataset \cite{nelson2008exploiting,paudice2018detection}, robust learning of the models in presence of corrupted data \cite{diakonikolas2018sever,zhu2019generalized}. However, defense against the data poisoning attack in online learning is not well-studied in the existing literature \cite{steinhardt2017certified,collingedefending,wang2019investigation}. \emph{Slab} defense \cite{steinhardt2017certified} is an effective data sanitization defense against many attacks \cite{wang2019investigation}. In this work, we propose an influence-based defense strategy in addition to the slab defense \cite{steinhardt2017certified} to minimize the impact of the poisoned data points on the learner's model.


    Influence function \cite{cook1982residuals} is a well-studied technique in robust statistics. In the recent past, machine learning researchers have used it to improve the model reliability \cite{schulam2019can}, to improve the model fairness \cite{wang2019repairing}, and to measure the group effects in the model prediction \cite{koh2019accuracy}. Also, the effect of a training point on a model prediction is efficiently estimated using the influence function without repeating the training process \cite{koh2017understanding}. 

    In this paper, we propose a method for defending against data poisoning attacks on online learning. Our contributions are as follows:
    \begin{itemize}
        \item We propose an influence-based defense against data poisoning attacks on online learning. We formulate the defense algorithm in two steps. First, we use the slab defense \cite{steinhardt2017certified} for initial data sanitization. Then, we apply our influence-based approach for minimizing the degradation caused by the poisoned data.
        \item We empirically investigate the performance of the proposed defense under multiple attacks and across multiple datasets.
    \end{itemize}

    The rest of the paper is organized as follows. Section \ref{sec:relatedworks} presents a brief survey of the related work. The threat model and system assumptions are discussed in Section \ref{sec:threat_model}. The necessary prerequisites are discussed in Section \ref{sec:prelim}, while Section \ref{sec:Influence_Defense} discusses our proposed Influence based defense for online learning. In Section \ref{sec:experiments}, we present the experiment results and observations. Finally, we conclude the paper in Section \ref{sec:conclusion}.

\section{Related Work} \label{sec:relatedworks}
    Data poisoning attacks are a serious threat to machine learning applications. There are extensive studies that have proposed attacks and defenses in the past. Biggio et al. \cite{biggio2012poisoning} have discussed data poisoning attack on Support Vector Machines (SVM). Mei et al. \cite{mei2015using} presented data input poisoning attack as a bilevel optimization problem and showed its effectiveness on SVM, logistic regression and linear regression. Koh et al. \cite{koh2017understanding} studied the use of influence functions for tracing a model's prediction through the learning algorithm and back to its training data. They further used this method for mounting a data input poisoning attack. Similarly, defending a machine learning model against such attacks is also an active and extensively studied area. Paudice et al. \cite{paudice2018detection} presented a defense mechanism to mitigate the effect of the poisoning attacks based on outlier detection. In a different work, Paudice et al. \cite{paudice2018label} have described a label flipping attack, a special case of data poisoning, and methods to mitigate such attacks. 
    
    Most of the research in this area considers offline learning, and less work has been reported on the attacks and defenses for the online setting. Wang and Chaudhuri \cite{wang2018data} presented an attack strategy, formulated as an optimization problem, for semi-online and fully-online settings. For their proposed attack, they considered a white-box adversary, where the attacker knows the entire data stream, the model class, the training algorithm and the deployed defenses. More recently, Wang et al. \cite{wang2019investigation} have studied the various defenses that can help mitigating the effect of data input poisoning on online learning. Their study showed that the \emph{Slab} defense is an effective method, and weaker threat models can result in fairly powerful attacks. Zhang et al. \cite{zhang2019online} formulated an online data poisoning attack as a stochastic optimal control problem where the attacker has no knowledge about the future training data and the distribution of data. Collinge et al. \cite{collingedefending} discussed an attack strategy against online learning classifiers in their work. They also discussed a defense strategy that analyzes the impact of a data point on the learning process and rejects suspicious inputs if their impact on the model is more than a threshold. An extra level of defense is proposed by training machine learning classifiers with different learning rates.

    Outlier detection is a common defense strategy, wherein the objective is to remove the training points that substantially deviate from the normal \cite{hodge2004survey}. Such approaches are well suited for removing typical noises in the data, but not for removing certain well-crafted adversarial noises. Also, these approaches cannot remove all the poisoned points from the data stream. Steinhardt et al \cite{steinhardt2017certified} in their work presented certified data sanitization defenses, where they considered two classes of defense: fixed defense and data-dependent defense with distinct outlier detection rules. Paudice et al \cite{paudice2018detection} introduced a defense strategy based on K-Nearest Neighbor (K-NN) that requires a trusted dataset. The requirement of a trusted dataset makes their approach unrealistic and impractical. Moreover, approaches that rely on a trusted dataset are vulnerable to the adversaries who can tamper it.

    Recent works have proposed attacks and defenses using techniques from robust statistics \cite{koh2017understanding}. Kang et al \cite{kang2019testing} studied methods for evaluating the defense strategies against unforeseen attacks. They have empirically shown that the robustness of a model based on the evaluation against single type of attack is not enough to provide information about the model robustness. Koh et al \cite{koh2018stronger} studied the effectiveness of data-dependent defenses ($L2$ and $Slab$) against certain attacks, for example, influence, min-max and KKT. They observed that the data-dependent defenses ($L2$ and $Slab$) perform better against min-max attack with MNIST and Dogfish datasets, but not with Enron and IMDB datasets. Zhu et al. \cite{zhu2019generalized} proposed a method that uses a generalization of minimum distance 
    functionals, which projects the poisoned data distribution onto a given set of well-behaved distributions. In this work, we combine the $Slab$ defense with the influence function \cite{koh2017understanding} for an efficient defense strategy. In rest of the paper we will refer our proposed approach as influence based approach.

    \section{Threat Model and Assumptions}\label{sec:threat_model}
    Attackers can be broadly classified into three types based on their capability and knowledge - offline, semi-online and fully-online \cite{wang2018data,wang2019investigation}. The objective of all three attacker models is to degrade the performance of the learner's model.

    \textbf{\emph{Offline Attacker.}} An offline attacker has access to the training data and can add poisoning data instances to the clean training data and she is oblivious to the streaming nature of the input. The attacker uses a method similar to \cite{biggio2012poisoning} for generating the poisoned data points. From an attacker's perspective, both the attack and the learning process happen in an offline fashion. 

    \textbf{\emph{Semi-online Attacker.}} A semi-online attacker can add or update the data instances at any position in the clean data stream. The resultant poisoned data is then used by the victim for training their ML model. Goal of the semi-online attacker is same as that of the offline attacker, that is, to degrade the performance of the final model. The attacker knows the entire clean data stream like an offline attacker, but the model is updated in an online manner.

    \textbf{\emph{Fully-online Attacker.}} A fully-online attacker knows the data stream up to the current time step and does not know about the future data instances. Hence, the fully-online attacker differs from the semi-online attacker as the latter has access to the entire data-stream. Moreover, the attacker can add or update the data points at pre-specified position only. The poisoned data resulted at specific time steps are then used by the learner for training. The attacker's objective is similar to an offline or semi-online attacker, that is, to degrade the victim model's performance, but over the entire time horizon.


    \textbf{\emph{Assumptions.}} For fair comparison with the baseline, we consider semi-online and fully-online attackers with attacker's capability similar to that of Wang et al. \cite{wang2019investigation}. We assume that the attacker's objective is to reduce the accuracy of the victim's model. In a semi-online attack, the entire clean data stream is known to the attacker and she can add a predetermined number of poisoned data points at any position in the stream. On the contrary, the capacity of the online attacker at a time step is limited to the data points received up to that time step. We assume a white-box setting, wherein the attacker know about the learner's model, the training algorithm, hyper-parameters and any defense deployed.
    
\section{Preliminaries}\label{sec:prelim}
    We develop a defense mechanism against data poisoning targeted at online binary classification task. In this section, we introduce data poisoning attacks against binary classifiers for online learning and then define the data poisoning attacker. Further, we define the data poisoning defender and discuss common defense strategies.

    \subsection{Data Poisoning Attacks}\label{sec:data_poisoning}
    Let us consider $\mathcal{D}$ = $\{(\bm{x}_{i},y_{i})\}_{i=1}^{n}$, where $\bm{x}_{i} \in \mathbb{R}^{d}$ is the $i^{th}$ input instance with $d$ dimensions and $y_{i} \in \{-1, +1\}$ is the corresponding binary classification label. A linear classifier can be learned using the objective function,
	\begin{equation}
	\begin{aligned}
	{\underset{f \in \mathcal{H}}{min}} \; \sum_{i=1}^{n} L(y_{i},f(\bm{x}_{i})) + \Omega(\theta)
	\label{eq:ObjFun}
	\end{aligned}
	\end{equation}
    where $f(\bm{x}_{i}) = \bm{\theta}^{T}\bm{x}_{i} + b$ is the decision function learned by minimizing the objective function (Equation \ref{eq:ObjFun}) on the data $\mathcal{D}$, $\Omega(\theta) = \frac{c}{2}||\theta||^{2}$ is a regularization parameter and $c$ is a constant, $\mathcal{H}$ is the hypothesis space, and $L$ is the loss function. For a given test instance $\bm{x}_{i}$, $\hat{y}_{i}= sgn(f(\bm{x}_{i}))$ is the predicted label. Additionally, we formalize the data poisoning attack as follows:

    \begin{definition}
        Data poisoning attacker takes as input a feasible data set $\mathcal{D}$, an initial parameter of the model $\theta_{0}$, the budget $K$, and outputs a poisoned version of the data samples $\mathcal{D}_{p}$ with at most $K$ samples are inserted or updated. The model parameter $\theta^{*}$ learned thereafter results in maximum error rate on test data $\mathcal{D}_{test}$.
    \end{definition}

    \subsection{Online Learning}
        In an online setting, arrival order of the data instances matter. Let the sequentially arrived data instances be $\{(x_{0},y_{0}), \dots, (x_{t},y_{t})\}$ and the model parameter $\theta_{t}$ at time $t$ is updated iteratively to $\theta_{t+1}$ based on the instance $(x_{t},y_{t})$. Online gradient descent (OGD) is used for updating $\theta_{t}$ at time $t$ with $\eta_{t}$ as the learning rate, $\theta_{0}$ as the initial model parameter \cite{wang2018data}. Therefore, the update at time $t$ is computed as:

        \begin{equation}\label{eq:OGD}
            \theta_{t+1} = \theta_{t} - \eta_{t} (\nabla L (\theta_{t},(x_{t},y_{t})) + \nabla \Omega(\theta_{t}))
        \end{equation}
        where $L$ is a convex loss function. Also, regularization parameter $\Omega(\theta) = \frac{c}{2}||\theta||^{2}$, where $c$ is a constant.

    \subsection{Data Poisoning Attack on Online Learning}
        An optimization problem from an attacker's perspective for a data stream is formulated in \cite{wang2018data}. Let $\mathcal{D}$ be a data stream and ($x_t, y_t$) be the data instance at time $t$. As the only difference between fully-online and semi-online is when attacker's objective is evaluated, therefore we discuss only the semi-online for brevity. Also, based on the evaluation (fully or semi), poisoned data is added either at the end or at any location in the stream. 
    
        For a prediction task for an input $x \in \mathcal{X}$ to an output $y \in \mathcal{Y}$, let $\mathcal{F}$ be a feasible set, such that $\mathcal{F} \subseteq \mathcal{X} \times \mathcal{Y}$. The feasible set $\mathcal{F}$ is of bounded diameter on which each example ($x_t, y_t$) is projected on arrival. This is an assumed defense used by the learner for excluding trivial attacks that tries to modify the learning process by adding outlier examples with very high norm. The learning process only considers data points for training that are in $\mathcal{F}$. As we are considering binary classification, therefore $\mathcal{Y} = \{-1, +1\}$ \cite{steinhardt2017certified}, and the feasible set is $\mathcal{F} = [-1,1]^d \times \{-1, 1\}$ where $d$ is the overall number of features \cite{biggio2012poisoning}. According to \cite{wang2018data}, under this defense an attacker would poison data points that are in the feasible set. 

        Let us consider a semi-online setting where $\mathcal{D}_{train}$ be the input data stream and $T$ denotes the length of the data stream, then the attacker's optimal strategy is the solution to the following optimization problem:
        \begin{equation}
            \begin{aligned}
                & \underset{\mathcal{D} \in \mathcal{F}^{T}}{max} \; g(\theta_{T}) \\ & \text{subject to:} \;  | \mathcal{D} - \mathcal{D}_{train} | \le K, \\
                & \theta_{t} = \theta_{t-1} - \sum_{\tau=0}^{t-1} \eta_{\tau}\;(\nabla L (\theta_{\tau}, \mathcal{D}_{\tau}) + \nabla \Omega(\theta_{\tau})), \; 1\le t \le T
            \end{aligned}
        \end{equation}
        where, $\mathcal{D} - \mathcal{D}_{train}$ is the difference between the two data streams, $g$ is the attacker's objective function. The attacker can poison at most $K$ instances in the input data stream and $\theta_{t}$ updated at time $t$ using the Online Gradient Descent (OGD) method. The objective function $g(\theta_t)$ depends on a specific weight vector $w$, that is the learned classifier and, the attack setting, for example, in fully-online setting, the objective function $g(\theta_{T})$ becomes $\sum_{t=0}^{T} g(\theta_{t})$. The attack can be done in two ways: (i) the attacker either manipulates the data in the input training data, or (ii) crafts the poisoned data and inserts it into the clean training data.
    
    \subsubsection{Targeted Attack Strategies} \label{sec:target}
        In our work, we evaluate the proposed defense strategy against the following data poisoning attacks:

        \textbf{\emph{Simplistic Attack} \cite{wang2018data,wang2019investigation}}. Let $\theta_{0}$ be the initial model and $\theta^{*}$ be the attacker's target. In this attack, some data points are appended to the clean data stream $\mathcal{D}_{c}$ which results in $\mathcal{D}_{p}$, further $\mathcal{D}_{p}$ is used for training the model $\theta^{p}$. The attack is considered to be successful when $\mathcal{D}_{p}$ satisfies the following three properties:

        \begin{enumerate}
            \item $|| \theta^{*} - \theta^{p}|| \le \epsilon$ where, $\epsilon$ is the tolerance parameter,
            \item at most $K$ data points are inserted into $\mathcal{D}_{c}$,
            \item the data points lie in the feasible set $\mathcal{F}$, where $\mathcal{F} \subseteq \mathcal{X} \times \mathcal{Y}$. 
        \end{enumerate}
        The attack projects the generated data sample onto $\mathcal{F}$ when it falls out of the feasible region.

        \textbf{\emph{Online attack \cite{collingedefending}}}. The attacker's objective is to change the state of the existing model by injecting the poisoned data points into the stream, thus degrading the system's performance. Let $g$ be the attacker's objective function at the iteration $t$ that is evaluated on the target dataset $D_c$. Also let attacker inject a poisoned data point $(x_t, y_t) \in \mathcal{F}$, such that:
        \begin{equation}
            \textbf{$x_t$}^{*} \in \underset{\textbf{$x_t$} \in \mathcal{F}}{argmax} \; g(\mathcal{D}_{c}, \theta_{t+1})
            \label{eq:onlineattack}
        \end{equation}
        where, $g$ is evaluated at iteration $t+1$ with the parameters of the classifier. For the learning rate $\eta$ and loss function $\mathcal{L}$, defined by the defender, the parameters are computed using the OGD as $\theta_{t+1} = \theta_t - \eta \bigtriangledown_{\theta_t} \mathcal{L}(x_t, \theta_t)$. Equation \ref{eq:onlineattack} can be solved using gradient ascent:
        \begin{equation}
            x_{t_{n+1}} = x_{t_n} + \alpha \bigtriangledown_{x_{t_n}} g(D_c, \theta_{t+1})
        \end{equation}
        where, $\alpha$ is the learning rate of the attacker in Equation \ref{eq:onlineattack}. The attack is mounted by first training the learning algorithm using the whole training set for some fixed number of iterations. After that, for each poisoning point a random subset of the training data is used, along with the poisoned points created in the previous iterations. The attack algorithm uses gradient descent for one epoch for updating the state of the learning algorithm and computing the poisoned data points \cite{collingedefending}.

    \subsection{Defenses against Data Poisoning Attacks}

    \emph{Data Sanitization} \cite{cretu2008casting} is a common defense against the data poisoning attacks, wherein the anomalous training points are removed before training the model. The objective is to remove the instances that are very different from the clean data instances. The existing defenses differ from each other on the basis of how they define an instance to be anomalous \cite{koh2018stronger}. Some of the defenses related to our work are \emph{L2}, \emph{slab} and \emph{loss}. The \emph{L2} defense discards points that are far away from the corresponding centroid, whereas, the \emph{slab} defense first projects points onto the line between the two class centroids and then discards points that are far away from the corresponding centroid. Similarly, the \emph{loss} defense discards data instances that are not well fit by the model on the full dataset.

    \section{Influence-based Defense for Online Learning}\label{sec:Influence_Defense}	
    Koh et al \cite{koh2017understanding} show that the \emph{influence functions} give us an efficient approximation on how the model's predictions change if we did not have the training data point $\mathcal{D}_{i}$. As per their notations, let $L(\mathcal{D}, \theta)$ be the loss and $\frac{1}{n}\sum_{i=1}^{n} L(\mathcal{D}_{i}, \theta)$ be the empirical risk, which can be calculated by averaging the loss function on the training set. Also, let $\mathcal{D}_{i}$ be the training data point and $\theta$ be the model parameter. The empirical risk minimizer is given by $ \hat{\theta} \stackrel{\text { def }}{=} \arg \min _{\theta \in \Theta} \frac{1}{n} \sum_{i=1}^{n} L\left(\mathcal{D}_{i}, \theta\right)$ with an assumption that $L$ is twice-differentiable and strictly convex in $\theta$.

    \begin{algorithm}[ht]
        \caption{Influence-based Defense Algorithm}
        \label{algo:Inf_def}
        \begin{algorithmic}[1]
            \renewcommand{\algorithmicrequire}{\textbf{Input:}}
            \renewcommand{\algorithmicensure}{\textbf{Parameter:}}
            \Require {Poisoned Data Stream $D$ = $\{(x,y)\}_{1}^{n}$, Pre-train data $D_{pre}$, initial model $\theta_{0}$ learned over pre-train data}
            \Ensure {Gradient descent step size $\eta_{def}$, influence window size $w_{inf}$}
            \renewcommand{\algorithmicensure}{\textbf{Output:}} 
            \Ensure {Model parameter $\theta_{n}$}
           
            \State Compute $\mathcal{F}_{slab}(D_{pre})$ on pre-train data using Equation \ref{eq:slab}
            
            \For{$x \in D$} \Comment{Doing for each $x$ from $D$ at time $t$} 
                \State Filter $x_{t}$ based on $\mathcal{F}_{slab}$
                \If{$x_{t}$ lies into feasible set $\mathcal{F}_{slab}(D_{pre})$}
                    \State Pre-compute the Hessian using Equation \ref{eq:hessian}
                    \State Pre-compute gradient loss using Equation \ref{eq:grad_loss}
                    \State $Inf(x_{t}) \leftarrow$ Compute the Influence using Equation \ref{eq:influence}
                    \State $Inf_{thres} \leftarrow$ Average of previous $w_{inf}$ influences
                    \If {$Inf(x_{t}) \ge Inf_{thres}$}
                        \State $x_{t}^{*}$ = $\underset{x}{\textbf{\emph{minimize}}} \; Inf(x)$ using Equation \ref{eq:grad_influence} 
                    	\If{$Inf(x_{t}) \ge Inf(x_{t}^{*})$}
                        	\State $x_{t} \leftarrow x_{t}^{*}$
                    	\EndIf
                   	\EndIf

                    \State $\theta_{t} \leftarrow $ Update $\theta_{t-1}$ using $x_{t}$
                \Else
                    \State continue without updating the $\theta_{t}$
                \EndIf
            \EndFor
            \State \textbf{return} $\theta_n$
           
        \end{algorithmic}
    \end{algorithm}
	
    Given a training point $\mathcal{D}_{i}$, the change in parameter can be defined formally as $\hat{\theta}_{-\mathcal{D}_{i}}-\hat{\theta}$ where,
    \begin{equation*}
        \hat{\theta}_{-\mathcal{D}_{i}} \stackrel{\text{ def }}{=} \arg \min _{\theta \in \Theta} \sum_{\substack{j=1 \dots n \\ \mathcal{D}_{j} \neq \mathcal{D}_{i}}} L\left(\mathcal{D}_{j}, \theta\right)
    \end{equation*}
    The idea is to compute the parameter change if $\mathcal{D}_{i}$ were upweighted by some small $\epsilon$, that is, $\mathcal{D}_{i}$ is multiplied by $1+\epsilon$. The influence of upweighting $\mathcal{D}_{i}$ on the parameters $\hat{\theta}$ is given by:

    \begin{equation}\label{eq:influence}
        \mathcal{I}_{(\mathcal{D}_i)} \overset{def}{=} 
        \frac{d \hat{\theta}_{\epsilon,\mathcal{D}_i}}{d\epsilon} \Bigr|_{\epsilon = 0} = -{H_{\hat{\theta}}^{-1}}\; \nabla_{\theta}
        {L}(\mathcal{D}_i, \hat{\theta}),
        \end{equation}
        where $H$ is the Hessian given by,
        \begin{equation}\label{eq:hessian}
        H_{\hat{\theta}} \stackrel{\text { def }}{=} \nabla^{2} R(\mathcal{D},\hat{\theta})=\frac{1}{n} \sum_{i=1}^{n} \nabla_{\theta}^{2} L(\mathcal{D}_{i}, \hat{\theta})
        \end{equation}
        
        \begin{equation}\label{eq:RTheta}
        R(\mathcal{D},\theta) \stackrel{\text { def }}{=} \frac{1}{n} \sum_{i=1}^{n} L\left(\mathcal{D}_{i}, \theta\right)
        \end{equation}
        
        \begin{equation}\label{eq:thetaCap}
        \hat{\theta}_{\epsilon, \mathcal{D}_i}=\arg \min _{\theta \in \Theta}\{R(\mathcal{D}, \theta)+\epsilon L(\mathcal{D}_i, \theta)\}
        \end{equation}
        
        Given a training point $\mathcal{D}_{i} = (\bm{x},y)$ $\in \mathcal{D}$, we first run a data-sanitization defense \cite{steinhardt2017certified} called $slab$ wherein $\mu_{y}$, $\mu_{-y}$ are class centroids and $s_{y}$ is threshold,
        \begin{equation}\label{eq:slab}
        \mathcal{F}_{\text {slab }} \stackrel{\text { def }}{=}\left\{(\bm{x}, y):\left|\left\langle \bm{x}-\mu_{y}, \mu_{y}-\mu_{-y}\right\rangle\right| \leq s_{y}\right\}
        \end{equation}
        Our objective is to reduce the degradation caused by the poisoned
        instances present in the feasible set. We achieve that by using influence functions (refer Equation \ref{eq:influence}). 
        
        Consider a generalized linear model function $h(.)$, parameterized by $\theta$. For example, the logistic regression classifier
        $$h_\theta(\bm{x})=\sigma(z)=\frac{1}{1+e^{-\theta^{T}\bm{x}}} = Pr(y=1 |~\bm{x}, \theta)$$
        where, $z = \theta^{T}\bm{x}$. Let the loss function be
        $$L(\theta)=\sum_{i=1}^n -\Big( y_i\log\sigma(z_i)+(1-y_i)\log(1-\sigma(z_i))\Big)$$
        Our focus is on the online setting where the weight vector is updated as in Equation \ref{eq:OGD}.
        The gradient of loss is given as
        \begin{equation}\label{eq:grad_loss}
        \begin{aligned}
        {\nabla}L(\mathcal{D}_{i},\theta) \; \;
        &\; = \; \frac{\partial L(\mathcal{D}_i,\theta)}{\partial \theta^T} \\
        & \; = \; -y_i\bm{x}_i(1-\sigma(z_i))+(1-y_i)\bm{x}_i\sigma(z_i) \\
        & \; =\; \bm{x}_i(\sigma(z_i)-y_i)
        \end{aligned}
        \end{equation}
        
        In order to achieve our objective, we first scalarize the influence function by taking the euclidean
        norm of it. If $\mathcal{S}$ is the scalarizing function, then
        $$
        \nabla \mathcal{S}(\mathcal{I}_{(\bm{x})})=\mathcal{J}( \mathcal{I}_{(\bm{x})})^{\top} \nabla \mathcal{S}(a) \; \Bigr|_{a = \mathcal{I}_{(\bm{x})}}
        $$
        where $\mathcal{J}$ is the jacobian matrix. $\nabla \mathcal{S}(a)$ is the gradient of $\mathcal{S}$ at the point $a = \mathcal{I}_{(\bm{x})} $.
        
        We then minimize the influence of training instances by applying gradient descent and perturbing the points.
        \begin{equation}\label{eq:min_inf}
        \bm{x}_{t}^{*} = \underset{\bm{x}}{\textbf{\emph{minimize}}} \; \mathcal{I}_{(\bm{x})}
        \end{equation} 
        \begin{equation}\label{eq:grad_influence}
        \nabla \mathcal{I}_{(\bm{x})}= - \nabla H^{-1}\left(\nabla_{\theta} L(\bm{x}, \theta)\right)
        \end{equation}
        
        \begin{equation}\label{eq:hessian_inf}
        H = {\nabla}^{2} L(\theta)=\sum_{i=1}^{n} \bm{x}_{i} \bm{x}_{i}^{T} \sigma\left(z_{i}\right)\left(1-\sigma\left(z_{i}\right)\right)
        \end{equation} 
        where hessian $H$ is a constant with respect to the incoming point. $\mathcal{I}_{(\bm{x})}$ decreases fastest if one goes from x in the direction of the negative gradient of $\mathcal{I}_{(\bm{x})}$ at $\bm{x}$ i.e., -$\nabla \mathcal{I}_{(\bm{x})}$. We may lose some information because we perturb the clean data as well.

        We formalize the influence-based defense as follows. \emph{Defender takes as input the data instances, then filters out the suspicious poisoning points and \textbf{minimizes the influence} of the remaining data points such that the learned model parameter $\hat{\theta}$ has the lowest error on a test data set $\mathcal{D}_{test}$}. The complete procedure is explained in the Algorithm \ref{algo:Inf_def}.

    \subsection{Effect of Slab and Influence Based Defense on Clean Data}
\begin{figure}[ht]
            \centering 
            \begin{subfigure}[b]{0.48\textwidth}
                \centering 
                \includegraphics[width=1\linewidth]{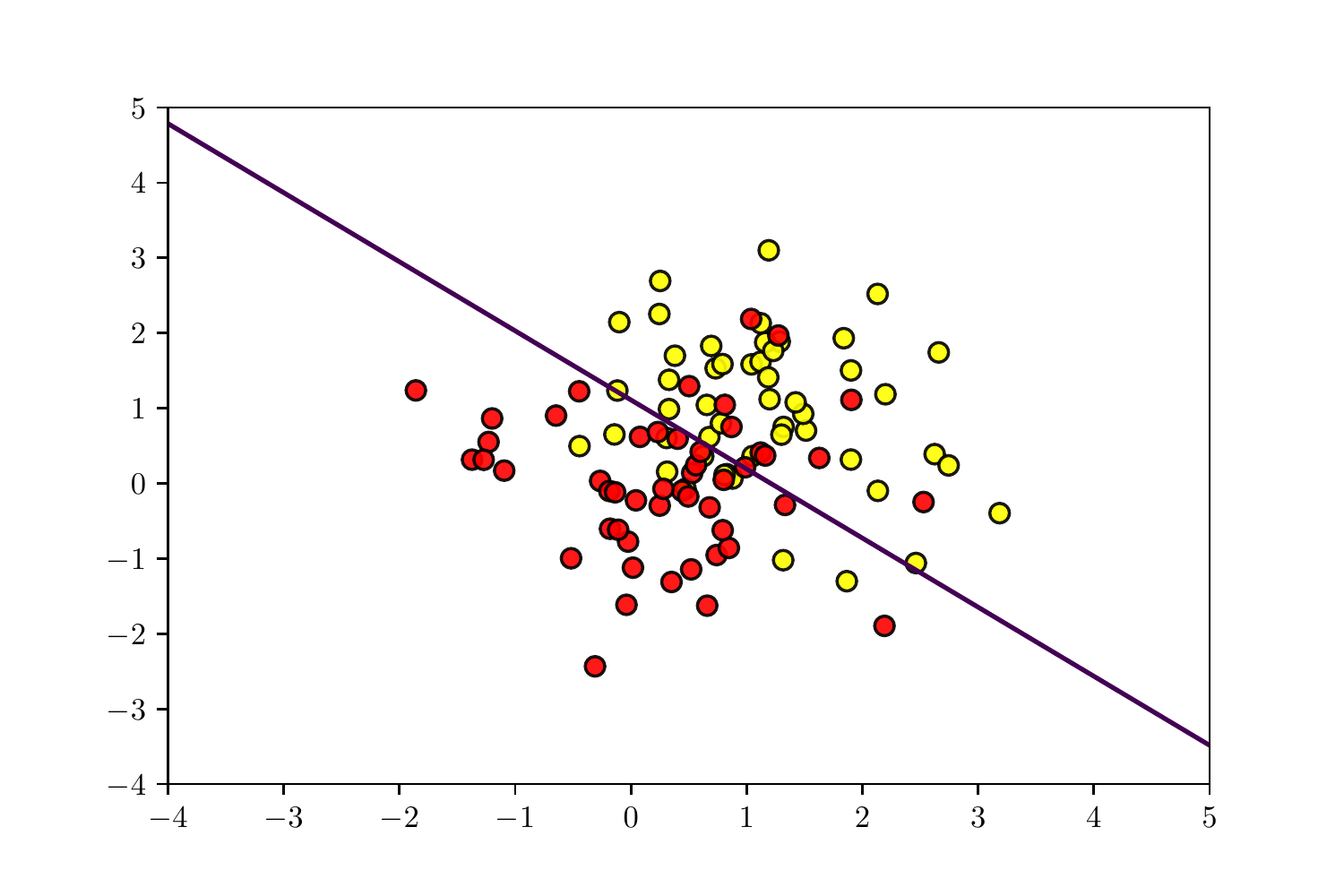} 
                \caption{}
                \label{fig:trainingdata}
            \end{subfigure}
            \begin{subfigure}[b]{0.48\textwidth}
                \centering \includegraphics[width=1\linewidth]{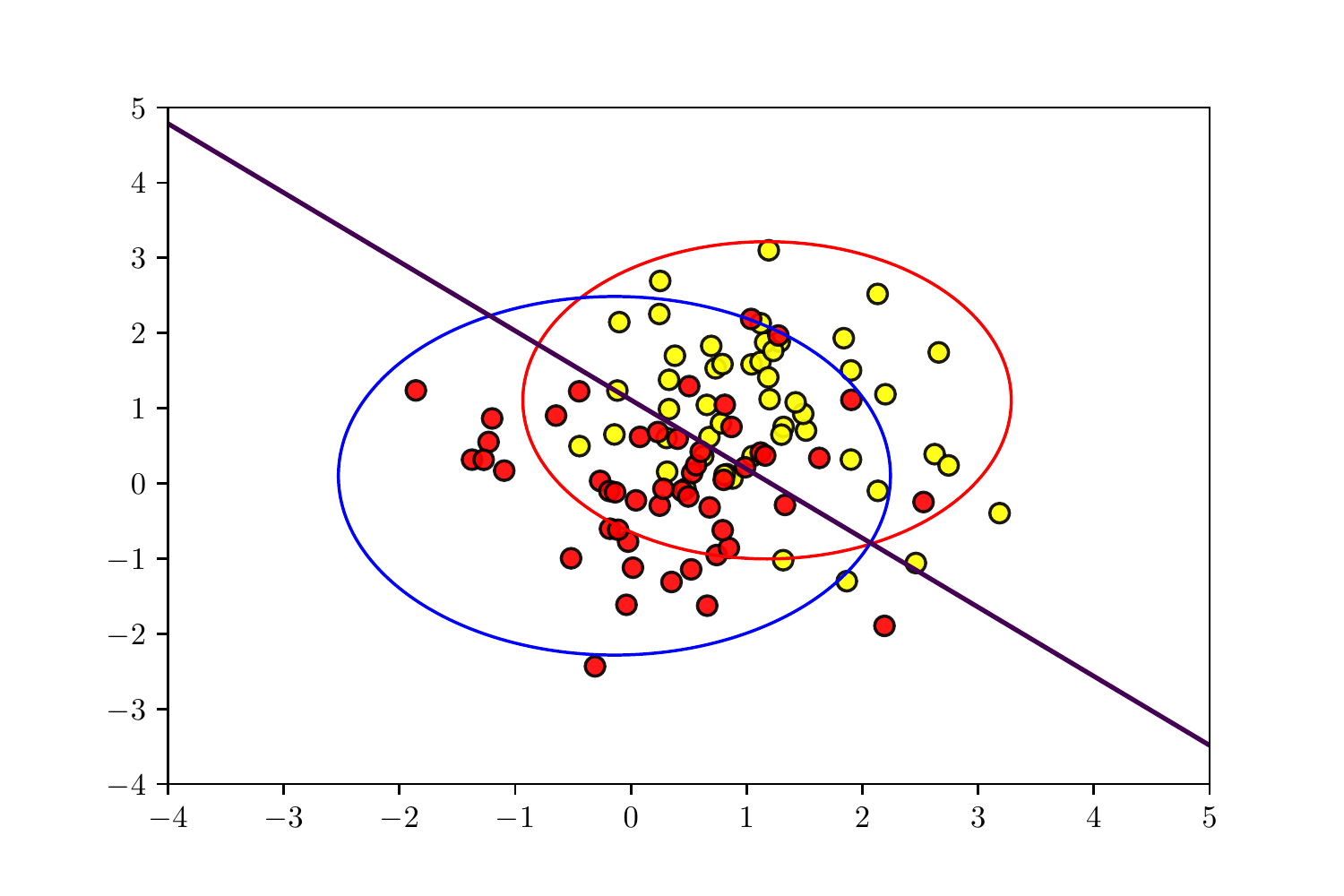} 
                \caption{}
                \label{fig:radius}
            \end{subfigure}

            \begin{subfigure}[b]{0.48\textwidth}
                \centering \includegraphics[width=1\linewidth]{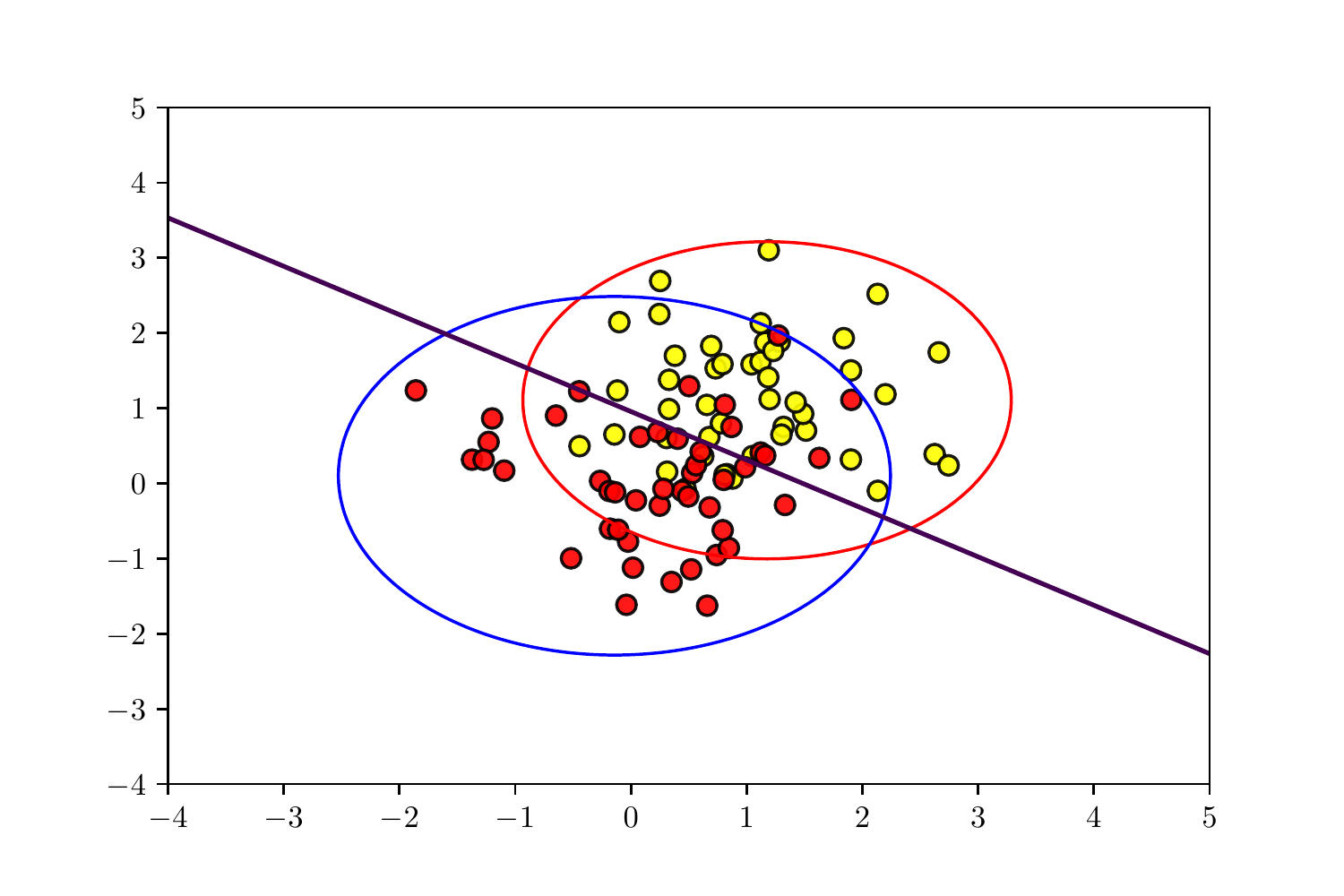} 
                \caption{}
                \label{fig:slab}
            \end{subfigure}
            \begin{subfigure}[b]{0.48\textwidth}
                \centering \includegraphics[width=1\linewidth]{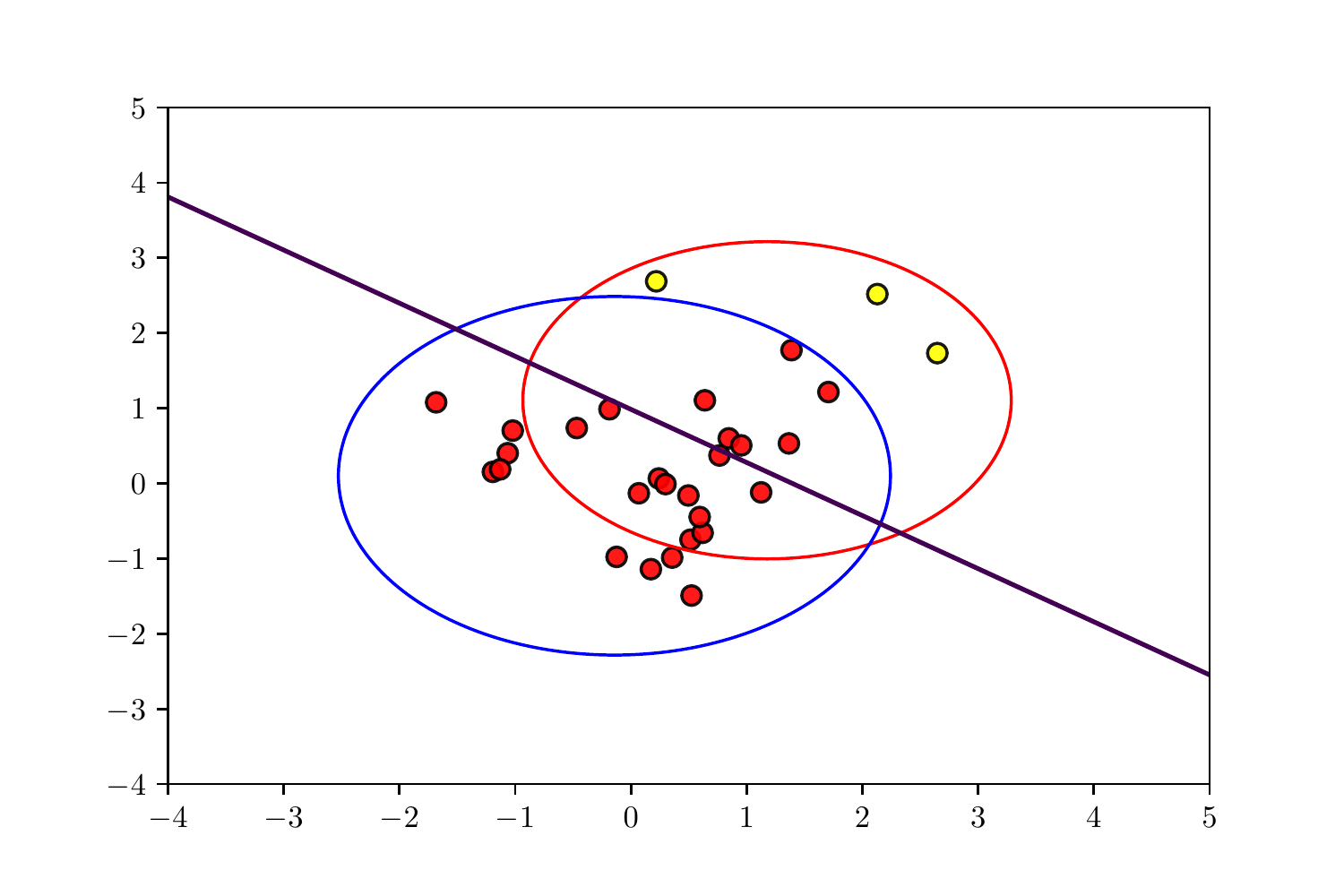} 
                \caption{}
                \label{fig:influence}
            \end{subfigure}
            \caption{Comparison of Slab and Influence based methods on synthetic data. Minimizing the influence sometimes affects the clean data points, which leads to information loss.}
            \label{fig:poisoning}
        \end{figure}

        To study the effect of Slab and Influence based approaches, we generated synthetic data using numpy's\footnote{https://numpy.org/} random package with different seed values for pre-train (size:50,seed:0), train (size:100,seed:1), validation (size:50,seed:2) and test data (size:50,seed:3). Figure \ref{fig:trainingdata} shows the plot for training data, where the decision boundary is computed on the training data. Figure \ref{fig:radius} and Figure \ref{fig:slab}, show the plot of training data with slab radius and training data after slab based sanitization. The decision boundary in Figure \ref{fig:slab} is computed on the sanitized data. Figure \ref{fig:influence} shows the plot of data points that changes after applying the influence-based method on sanitized data. The decision boundary in Figure \ref{fig:influence} is computed using both the changed and unchanged data after applying influence-based defense. The accuracy of the classifier on the training data is observed to be 76\%. After applying the slab defense the accuracy changes to 68\%, however the accuracy reduces to 66\% after applying influence based defense. This drop in accuracy is due to `change' of some of the clean data points with high influence score. 

        \begin{table}[ht]
            \centering
            \caption{Datasets used for comparing Slab and Influence based method.}
            \label{tab:datasets}
            \begin{tabular}{@{}lccccc@{}}
            \toprule
            \textbf{\#} & \textbf{Dataset}  & \textbf{Features} & \textbf{Pre-Train} & \textbf{Training} & \textbf{Test} \\ \midrule
            \textbf{D1}           & Australian        & 14                & 200                & 300               & 150           \\
            \textbf{D2}           & Banknote          & 4                 & 200                & 400               & 572           \\
            \textbf{D3}           & MNIST 1v7         & 50                & 8000               & 1000              & 2163          \\
            \textbf{D4}           & Spambase          & 57                & 2000               & 1000              & 1519          \\
            \textbf{D5}           & UCI Breast Cancer & 9                 & 100                & 400               & 100           \\
            \textbf{D6}           & Fashion MNIST (Bag versus Sandal)     & 50                & 8000               & 1000              & 1000          \\ \bottomrule
            \end{tabular}
        \end{table}
    
\section{Experiments}\label{sec:experiments}
    In this section we will briefly discuss the datasets that we have used, the test environment, experimental results and observations. 

    \textbf{\emph{Hardware and Software Configuration.}} We have carried out all our experiments on a machine having Intel Core i7-8550U CPU with 4 cores, 8 logical processors and a base speed of 2.0 GHz. The system has 16 GB DDR4 RAM and 500 GB HD with 230 GB free space. Also, the test system has L1, L2 and L3 cache of size 256 KB, 1 MB and 8 MB respectively. Our proposed defense is implemented using Python 3.6 with scikit-learn\footnote{https://scikit-learn.org/stable/supervised\_learning.html} implementation of the classifiers, and tested on 64-bit Windows 10 enterprise edition and Ubuntu 16.04 LTS version. 

    \begin{table}[ht]
        \centering
        \caption{Comparison of results on six datasets for 10\% poisoning budget and with different learning rates. Results, where one method is better than the other, are in bold. The datasets \textbf{D1} to \textbf{D6} are discussed in Table \ref{tab:datasets}. \textbf{Slab (\textit{S})} and \textbf{Influence-based (\textit{I})} defenses are compared against the Simplistic and Online attacks.}
        \label{tab:results}
        \begin{tabular}{@{}cccccccccc@{}}
        \toprule
        &  & \multicolumn{2}{c}{\textbf{LR: 0.01}} & \multicolumn{2}{c}{\textbf{LR: 0.05}} & \multicolumn{2}{c}{\textbf{LR: 0.09}} & \multicolumn{2}{c}{\textbf{LR: Optimal}} \\ \midrule
        &  & \textit{Simplistic} & \textit{Online} & \textit{Simplistic} & \textit{Online} & \textit{Simplistic} & \textit{Online} & \textit{Simplistic} & \textit{Online} \\
        \multirow{2}{*}{\textbf{D1}} & \textit{S} & 0.8736 & 0.8736 & 0.8736 & 0.8526 & 0.8789 & 0.7947 & 0.8578 & 0.5736 \\
        & \textit{I} & 0.8736 & 0.8736 & 0.8736 & \textbf{0.8578} & 0.8789 & \textbf{0.8052} & 0.8578 & \textbf{0.6211} \\
        \multirow{2}{*}{\textbf{D2}} & \textit{S} & 0.8391 & 0.8671 & \textbf{0.8287} & 0.8933 & \textbf{0.7639} & \textbf{0.9161} & 0.4423 & 0.9493 \\
        & \textit{I} & 0.8391 & 0.8671 & 0.806 & \textbf{0.8951} & 0.5804 & 0.8951 & 0.4423 & \textbf{0.9755} \\
        \multirow{2}{*}{\textbf{D3}} & \textit{S} & 0.9773 & 0.9884 & \textbf{0.9242} & \textbf{0.9917} & \textbf{0.8335} & 0.9801 & 0.6643 & 0.5839 \\
        & \textit{I} & \textbf{0.9787} & \textbf{0.9889} & 0.8974 & 0.9847 & 0.8220 & \textbf{0.9810} & \textbf{0.7753} & \textbf{0.9223} \\
        \multirow{2}{*}{\textbf{D4}} & \textit{S} & 0.8933 & 0.8531 & 0.6893 & 0.7544 & 0.6372 & \textbf{0.7281} & 0.6254 & \textbf{0.7182} \\
        & \textit{I} & \textbf{0.8940} & 0.8531 & \textbf{0.6899} & \textbf{0.7551} & \textbf{0.6458} & 0.7228 & \textbf{0.6445} & 0.6886 \\
        \multirow{2}{*}{\textbf{D5}} & \textit{S} & 0.92 & 0.88 & 0.94 & 0.53 & 0.94 & 0.52 & 0.88 & 0.49 \\
        & \textit{I} & 0.92 & 0.88 & 0.94 & 0.53 & 0.94 & \textbf{0.68} & 0.88 & \textbf{0.71} \\
        \multirow{2}{*}{\textbf{D6}} & \textit{S} & 0.986 & 0.922 & 0.985 & \textbf{0.767} & 0.988 & 0.697 & 0.988 & 0.981 \\
        & \textit{I} & 0.986 & \textbf{0.926} & \textbf{0.988} & 0.733 & 0.988 & \textbf{0.737} & 0.988 & 0.981 \\ \cmidrule(l){2-10} 
        \end{tabular}
    \end{table}

    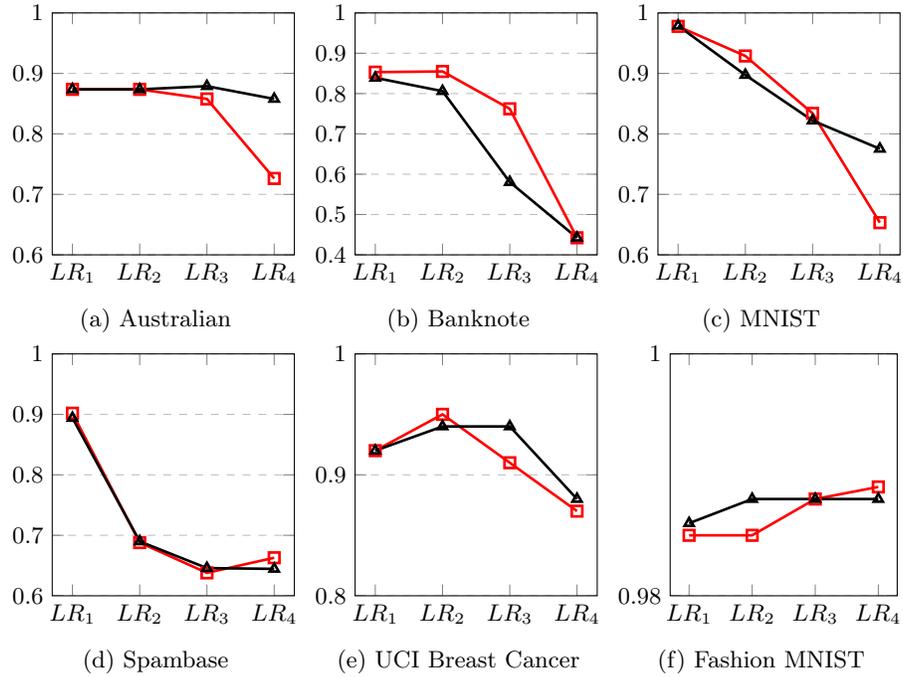
\begin{figure}[ht]
        \centering
        \begin{subfigure}[b]{0.33\textwidth}
            \begin{tikzpicture}
                \begin{axis}[ymin=0.6, ymax=1, xtick={0.01,0.05,0.09,0.13}, xticklabels={$LR_{1}$,$LR_2$,$LR_3$,$LR_4$}, ytick={0.6, 0.7, 0.8,0.9,1.0}, ymajorgrids=true, grid style=dashed, width=4.8cm, height=4.8cm]
                \addplot[color=red,mark=square,line width=1pt] coordinates {(0.01,0.8736)(0.05,0.8736)(0.09,0.8578)(0.13,0.7263)};
                \addplot[color=black,mark=triangle,line width=1pt] coordinates {(0.01,0.8736)(0.05,0.8736)(0.09,0.8789)(0.13,0.8578)};
                \end{axis}
            \end{tikzpicture}
            \caption{Australian}
        \end{subfigure}
        \begin{subfigure}[b]{0.33\textwidth}
            \begin{tikzpicture}
                \begin{axis}[ymin=0.4, ymax=1, xtick={0.01,0.05,0.09,0.13}, xticklabels={$LR_{1}$,$LR_2$,$LR_3$,$LR_4$}, ytick={0.4,0.5,0.6, 0.7,0.8,0.9,1.0}, ymajorgrids=true, grid style=dashed, width=4.8cm, height=4.8cm]
                \addplot[color=red,mark=square,line width=1pt] coordinates {(0.01,0.8531)(0.05,0.8549)(0.09,0.7622)(0.13,0.4423)};
                \addplot[color=black,mark=triangle,line width=1pt] coordinates {(0.01,0.8391)(0.05,0.806)(0.09,0.5804)(0.13,0.4423)};
                \end{axis}
            \end{tikzpicture}
            \caption{Banknote}
        \end{subfigure}
        \begin{subfigure}[b]{0.33\textwidth}
            \begin{tikzpicture}
                \begin{axis}[ymin=0.6, ymax=1, xtick={0.01,0.05,0.09,0.13}, xticklabels={$LR_{1}$,$LR_2$,$LR_3$,$LR_4$}, ytick={0.6, 0.7,0.8,0.9,1.0}, ymajorgrids=true, grid style=dashed, width=4.8cm, height=4.8cm]
                \addplot[color=red,mark=square,line width=1pt] coordinates {(0.01,0.9778)(0.05,0.9288)(0.09,0.8340)(0.13,0.6532)};
                \addplot[color=black,mark=triangle,line width=1pt] coordinates {(0.01,0.9787)(0.05,0.8974)(0.09,0.8220)(0.13,0.7753)};
                \end{axis}
            \end{tikzpicture}
            \caption{MNIST}
        \end{subfigure}

        \begin{subfigure}[b]{0.33\textwidth}
            \begin{tikzpicture}
                \begin{axis}[ymin=0.6, ymax=1, xtick={0.01,0.05,0.09,0.13}, xticklabels={$LR_{1}$,$LR_2$,$LR_3$,$LR_4$}, ytick={0.6, 0.7,0.8,0.9,1.0}, ymajorgrids=true, grid style=dashed, width=4.8cm, height=4.8cm]
                \addplot[color=red,mark=square,line width=1pt] coordinates {(0.01,0.9019)(0.05,0.688)(0.09,0.6379)(0.13,0.6629)};
                \addplot[color=black,mark=triangle,line width=1pt] coordinates {(0.01,0.8940)(0.05,0.6899)(0.09,0.6458)(0.13,0.6445)};
                \end{axis}
            \end{tikzpicture}
            \caption{Spambase}
        \end{subfigure}
        \begin{subfigure}[b]{0.33\textwidth}
            \begin{tikzpicture}
                \begin{axis}[ymin=0.8, ymax=1, xtick={0.01,0.05,0.09,0.13}, xticklabels={$LR_{1}$,$LR_2$,$LR_3$,$LR_4$}, ytick={0.8,0.9,1.0}, ymajorgrids=true, grid style=dashed, width=4.8cm, height=4.8cm]
                \addplot[color=red,mark=square,line width=1pt] coordinates {(0.01,0.92)(0.05,0.95)(0.09,0.91)(0.13,0.87)};
                \addplot[color=black,mark=triangle,line width=1pt] coordinates {(0.01,0.92)(0.05,0.94)(0.09,0.94)(0.13,0.88)};
                \end{axis}
            \end{tikzpicture}
            \caption{UCI Breast Cancer}
        \end{subfigure}
        \begin{subfigure}[b]{0.33\textwidth}
            \begin{tikzpicture}
                \begin{axis}[ymin=0.98, ymax=1, xtick={0.01,0.05,0.09,0.13}, xticklabels={$LR_{1}$,$LR_2$,$LR_3$,$LR_4$}, ytick={0.98,1.0}, ymajorgrids=true, grid style=dashed, width=4.6cm, height=4.8cm]
                \addplot[color=red,mark=square,line width=1pt] coordinates {(0.01,0.985)(0.05,0.985)(0.09,0.988)(0.13,0.989)};
                \addplot[color=black,mark=triangle,line width=1pt] coordinates {(0.01,0.986)(0.05,0.988)(0.09,0.988)(0.13,0.988)};
                \end{axis}
            \end{tikzpicture}
            \caption{Fashion MNIST}
        \end{subfigure}
        \caption{Accuracy versus Learning Rate comparison for classifier with and without defense for \textbf{simplistic attack}. The \ref{L2} refers to classifier with Influence based defense and \ref{L1} refers to classifier without defense. $LR_{1}$,$LR_2$,$LR_3$ and $LR_4$ refers to 0.01, 0.05, 0.09 and optimal learning rates.}
        \label{fig:simplistic}
    \end{figure}

    \begin{figure}[ht]
        \centering
        \begin{subfigure}[b]{0.33\textwidth}
            \begin{tikzpicture}
                \begin{axis}[ymin=0.4, ymax=1, xtick={0.01,0.05,0.09,0.13}, xticklabels={$LR_{1}$,$LR_2$,$LR_3$,$LR_4$}, ytick={0.4,0.5,0.6, 0.7, 0.8,0.9,1.0}, ymajorgrids=true, grid style=dashed, width=4.8cm, height=4.8cm]
                \addplot[color=red,mark=square,line width=1pt] coordinates {(0.01,0.8473)(0.05,0.5526)(0.09,0.5263)(0.13,0.6421)}; \label{L1}
                \addplot[color=black,mark=triangle,line width=1pt] coordinates {(0.01,0.8736)(0.05,0.8578)(0.09,0.8052)(0.13,0.6210)};\label{L2}
                \end{axis}
            \end{tikzpicture}
            \caption{Australian}
        \end{subfigure}
        \begin{subfigure}[b]{0.33\textwidth}
            \begin{tikzpicture}
                \begin{axis}[ymin=0.5, ymax=1, xtick={0.01,0.05,0.09,0.13}, xticklabels={$LR_{1}$,$LR_2$,$LR_3$,$LR_4$}, ytick={0.5,0.6, 0.7, 0.8,0.9,1.0}, ymajorgrids=true, grid style=dashed, width=4.8cm, height=4.8cm]
                \addplot[color=red,mark=square,line width=1pt] coordinates {(0.01,0.7727)(0.05,0.6573)(0.09,0.5769)(0.13,0.5576)};
                \addplot[color=black,mark=triangle,line width=1pt] coordinates {(0.01,0.8671)(0.05,0.8951)(0.09,0.8951)(0.13,0.9755)};
                \end{axis}
            \end{tikzpicture}
            \caption{Banknote}
        \end{subfigure}
        \begin{subfigure}[b]{0.33\textwidth}
            \begin{tikzpicture}
                \begin{axis}[ymin=0.4, ymax=1, xtick={0.01,0.05,0.09,0.13}, xticklabels={$LR_{1}$,$LR_2$,$LR_3$,$LR_4$}, ytick={0.4,0.5,0.6, 0.7, 0.8,0.9,1.0}, ymajorgrids=true, grid style=dashed, width=4.8cm, height=4.8cm]
                \addplot[color=red,mark=square,line width=1pt] coordinates {(0.01,0.9870)(0.05,0.4877)(0.09,0.4845)(0.13,0.4845)};
                \addplot[color=black,mark=triangle,line width=1pt] coordinates {(0.01,0.9889)(0.05,0.9847)(0.09,0.9810)(0.13,0.9223)};
                \end{axis}
            \end{tikzpicture}
            \caption{MNIST}
        \end{subfigure}

        \begin{subfigure}[b]{0.33\textwidth}
            \begin{tikzpicture}
                \begin{axis}[ymin=0.6, ymax=1, xtick={0.01,0.05,0.09,0.13}, xticklabels={$LR_{1}$,$LR_2$,$LR_3$,$LR_4$}, ytick={0.6, 0.7, 0.8,0.9,1.0}, ymajorgrids=true, grid style=dashed, width=4.8cm, height=4.8cm]
                \addplot[color=red,mark=square,line width=1pt] coordinates {(0.01,0.8077)(0.05,0.7564)(0.09,0.7175)(0.13,0.6787)};
                \addplot[color=black,mark=triangle,line width=1pt] coordinates {(0.01,0.8531)(0.05,0.7551)(0.09,0.7228)(0.13,0.6886)};
                \end{axis}
            \end{tikzpicture}
            \caption{Spambase}
        \end{subfigure}
        \begin{subfigure}[b]{0.33\textwidth}
            \begin{tikzpicture}
                \begin{axis}[ymin=0.3, ymax=1, xtick={0.01,0.05,0.09,0.13}, xticklabels={$LR_{1}$,$LR_2$,$LR_3$,$LR_4$}, ytick={0.3,0.4,0.5,0.6, 0.7, 0.8,0.9,1.0}, ymajorgrids=true, grid style=dashed, width=4.8cm, height=4.8cm]
                \addplot[color=red,mark=square,line width=1pt] coordinates {(0.01,0.6)(0.05,0.44)(0.09,0.38)(0.13,0.43)};
                \addplot[color=black,mark=triangle,line width=1pt] coordinates {(0.01,0.88)(0.05,0.53)(0.09,0.68)(0.13,0.71)};
                \end{axis}
            \end{tikzpicture}
            \caption{UCI Breast Cancer}
        \end{subfigure}
        \begin{subfigure}[b]{0.33\textwidth}
            \begin{tikzpicture}
                \begin{axis}[ymin=0.3, ymax=1, xtick={0.01,0.05,0.09,0.13}, xticklabels={$LR_{1}$,$LR_2$,$LR_3$,$LR_4$}, ytick={0.3,0.4,0.5,0.6, 0.7, 0.8,0.9,1.0}, ymajorgrids=true, grid style=dashed, width=4.8cm, height=4.8cm]
                \addplot[color=red,mark=square,line width=1pt] coordinates {(0.01,0.424)(0.05,0.45)(0.09,0.501)(0.13,0.455)};
                \addplot[color=black,mark=triangle,line width=1pt] coordinates {(0.01,0.926)(0.05,0.733)(0.09,0.737)(0.13,0.981)};
                \end{axis}
            \end{tikzpicture}
            \caption{Fashion MNIST}
        \end{subfigure}
        \caption{Accuracy versus Learning Rate comparison for classifier with and without defense for \textbf{online attack}. The \ref{L2} refers to classifier with Influence based defense and \ref{L1} refers to classifier without defense. $LR_{1}$,$LR_2$,$LR_3$ and $LR_4$ refers to 0.01, 0.05, 0.09 and optimal learning rates.}
        \label{fig:online}
    \end{figure}
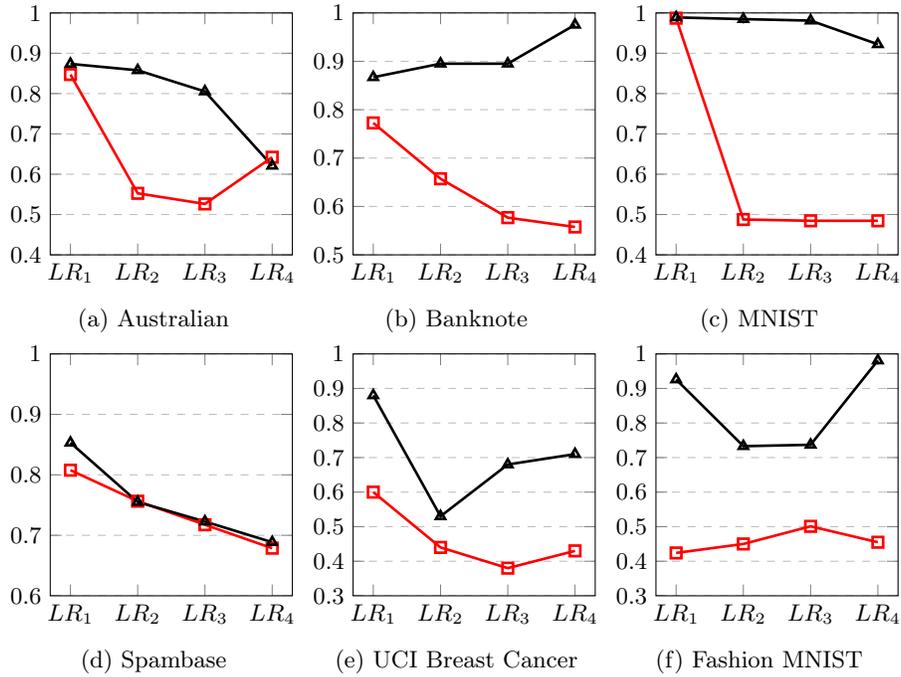

    \textbf{\emph{Datasets.}} We consider 6 datasets for our experiments. Table \ref{tab:datasets} shows the list of the datasets, number of features used for training the model and the distribution of the data for initialization, training and testing. The poisoning budget computation is relative to the training data. We validate our results on six datasets, three of which are common in \cite{wang2019investigation}.
        
    \textbf{\emph{Baseline Attacks.}} We consider two existing poisoning attacks on online learning for evaluating our defense strategy: a) \emph{simplistic attack} \cite{wang2018data}, and b) \emph{online attack} \cite{collingedefending} (refer to Section \ref{sec:target}). We have used the constant learning rates of 0.01 and 0.05 as per \cite{wang2019investigation} and an additional constant learning rate of 0.09. Moreover, we have also used the optimal learning rate provided by the scikit-learn to assess the performance of Slab and our influence based method. 
            
    \textbf{\emph{Defenses.}} Our defense is in conjunction with the slab defense. The 
    influence window size ($w_{inf}$) in our defense algorithm is empirically found using grid search. The poisoned points are generated by the adversary using the baseline attacks. The defender first uses slab defense for data sanitization in the incoming data stream. She then minimizes the influence of the data point based on its impact on the model. The defense mechanism is effective when the classifier has higher score than the classifier under baseline attacks.
            
    \textbf{\emph{Results and Discussion.}} Table \ref{tab:results} shows the effectiveness of Slab (\textit{S}) and Influence based (\textit{I}) defenses against Simplistic and Online attacks. We have performed experiments on six different datasets .We have considered three constant learning rates of 0.01, 0.05 and 0.09 respectively. In addition, we have also compared the two defenses for an optimal learning rate. It can be observed that on an average the performance of slab and influence based methods are nearly same for constant learning rate. On the contrary, for an optimal learning rate, the influence based method has far better result than slab for most of the datasets. For example, Influence defense is more than 9\% accurate than Slab for simplistic attack and more than 30\% accurate than online attack for MNIST dataset. 

    As per our observation, the simplistic attack is apparently more powerful than the online attack. This is due to the fact that simplistic attack inserts poisoned examples which lie in the feasible set $\mathcal{F}$, whereas the online attack of \cite{collingedefending} does not ensure that. We found that the number of poisoned examples kept after the Slab defense is almost equal to the attacker's budget against the simplistic attack. However, comparatively few poisoned examples (in some cases zero) are kept against the online attack. Comparatively the low accuracy of influence based method in some cases could be because of the change in the influence of clean samples. Also, the percentage of clean data samples that remain after slab defense for datasets \textbf{D1} to \textbf{D6} are 43.33, 76.5, 60.8, 69.8, 97.75, 90.90 respectively, for both the attacks. The percentage difference is due to the data-dependent nature of the slab defense \cite{wang2019investigation}. Table \ref{tab:results_5_budget} and Table \ref{tab:results_15_budget} show comparative results for 5\% and 15\% poisoning budget (Section \ref{sec:appresult}).

    Figure \ref{fig:simplistic} shows the comparison of accuracy of the classifier with and without influence based defense for simplistic attack for varying learning rates. Similarly, Figure \ref{fig:online} shows the change in classifier's accuracy with and without influence defense for the online attack. The poison budget is kept 10\% and learning rate is changed from fixed (0.01, 0.05, 0.09) to optimal. One common observation is that the accuracy of the classifier degrades more with optimal learning rate and in absence of a defense. For online attack of \cite{collingedefending}, our proposed works well to improve the accuracy in most cases. On the contrary, for simplistic attack our defense has relatively lower performance as the poisoned data points are added in the feasible set only, that is irrespective of their influence.

\section{Conclusion}\label{sec:conclusion}
    In this paper, we formulated a defense algorithm that compliments the slab defense with the influence function such that the degradation caused by the poisoned data is minimized. We studied the performance of our defense against different attacks and across multiple datasets. Further, we validated our experiments with simplistic and online attacks of \cite{wang2019investigation,collingedefending}. We have also demonstrated the performance degradation of the classifier with and without defense. One of the trade-offs from the defender's side is that the objective function to minimize the influence sometimes affects the clean data points, which leads to information loss.

\bibliographystyle{splncs04}
\bibliography{onlinedefense}

\appendix

\section{Additional Experiment Results} \label{sec:appresult}
    Table \ref{tab:results_5_budget} and Table \ref{tab:results_15_budget} shows the comparison of Slab and Influence based approaches on 5\% and 15\% poison budget for varying learning rates. It can be observed that the findings for these two additional poison budget is similar to that of 10\% budget.

    \begin{table}[ht]
        \centering
        \caption{The comparative results on six datasets for 5\% poisoning budget across different learning rates. Results, where one method is better than the other, are in bold. The datasets \textbf{D1} to \textbf{D6} are discussed in Table \ref{tab:datasets}. Both the defenses are studied against the simplistic and online attacks.}
        \label{tab:results_5_budget}
        \begin{tabular}{@{}cccccccccc@{}}
            \toprule
            & \textbf{}  & \multicolumn{2}{c}{\textbf{LR: 0.01}} & \multicolumn{2}{c}{\textbf{LR: 0.05}} & \multicolumn{2}{c}{\textbf{LR: 0.09}} & \multicolumn{2}{c}{\textbf{LR: Optimal}} \\ \midrule
            &            & \textit{Simplistic} & \textit{Online} & \textit{Simplistic} & \textit{Online} & \textit{Simplistic} & \textit{Online} & \textit{Simplistic}   & \textit{Online}  \\
            \multirow{2}{*}{\textbf{D1}} & \textit{S} & 0.8737              & 0.8737          & 0.8737              & 0.8526          & 0.8789              & 0.7947          & \textbf{0.8632}       & 0.5737           \\
            & \textit{I} & 0.8737              & 0.8737          & \textbf{0.8789}     & \textbf{0.8579} & 0.8789              & \textbf{0.8053} & 0.8579                & \textbf{0.6211}  \\
            \multirow{2}{*}{\textbf{D2}} & \textit{S} & 0.8566              & 0.8671          & \textbf{0.8776}     & 0.8934          & \textbf{0.8636}     & \textbf{0.9161} & 0.4423                & 0.9493           \\
            & \textit{I} & 0.8566              & 0.8671          & 0.8584              & \textbf{0.8951} & 0.6206              & 0.8951          & 0.4423                & \textbf{0.9755}  \\
            \multirow{2}{*}{\textbf{D3}} & \textit{S} & 0.9838              & 0.9884          & \textbf{0.9718}     & \textbf{0.9917} & \textbf{0.9362}     & 0.9801          & 0.7661                & 0.5839           \\
            & \textit{I} & \textbf{0.9843}     & \textbf{0.9889} & 0.9575              & 0.9847          & 0.9038              & \textbf{0.9810} & \textbf{0.8160}       & \textbf{0.9223}  \\
            \multirow{2}{*}{\textbf{D4}} & \textit{S} & 0.9118              & 0.8532          & \textbf{0.8038}     & 0.7544          & \textbf{0.7169}     & \textbf{0.7281} & 0.6333                & \textbf{0.7182}  \\
            & \textit{I} & 0.9118              & 0.8532          & 0.8032              & \textbf{0.7551} & 0.7123              & 0.7228          & \textbf{0.6498}       & 0.6886           \\
            \multirow{2}{*}{\textbf{D5}} & \textit{S} & 0.9200              & 0.8800          & 0.9400              & 0.5300          & \textbf{0.9500}     & 0.5200          & 0.8800                & 0.4900           \\
            & \textit{I} & 0.9200              & 0.8800          & 0.9400              & 0.5300          & 0.9400              & \textbf{0.6800} & 0.8800                & \textbf{0.7100}  \\
            \multirow{2}{*}{\textbf{D6}} & \textit{S} & \textbf{0.9880}     & 0.9220          & 0.9890              & \textbf{0.7670} & \textbf{0.9900}     & \textbf{0.9920} & 0.9880                & 0.9810           \\
            & \textit{I} & 0.9870              & \textbf{0.9260} & \textbf{0.9910}     & 0.7330          & 0.6970              & 0.7370          & 0.9880                & 0.9810           \\ \cmidrule(l){2-10} 
        \end{tabular}
    \end{table}

    \begin{table}[ht]
        \centering
        \caption{The comparative results on six datasets for 15\% poisoning budget across different learning rates. Results, where one method is better than the other, are in bold. The datasets \textbf{D1} to \textbf{D6} are discussed in Table \ref{tab:datasets}. Both the defenses are studied against the simplistic and online attacks.}
        \label{tab:results_15_budget}
        \begin{tabular}{@{}cccccccccc@{}}
            \toprule
            & \textbf{}  & \multicolumn{2}{c}{\textbf{LR: 0.01}}                                         & \multicolumn{2}{c}{\textbf{LR: 0.05}}                                         & \multicolumn{2}{c}{\textbf{LR: 0.09}}                                         & \multicolumn{2}{c}{\textbf{LR: Optimal}}                                      \\ \midrule
            &            & \multicolumn{1}{c}{\textit{Simplistic}} & \multicolumn{1}{c}{\textit{Online}} & \multicolumn{1}{c}{\textit{Simplistic}} & \multicolumn{1}{c}{\textit{Online}} & \multicolumn{1}{c}{\textit{Simplistic}} & \multicolumn{1}{c}{\textit{Online}} & \multicolumn{1}{c}{\textit{Simplistic}} & \multicolumn{1}{c}{\textit{Online}} \\
            \multirow{2}{*}{\textbf{D1}} & \textit{S} & 0.8684                                  & 0.8737                              & 0.8789                                  & 0.8526                              & 0.8684                                  & 0.7947                              & 0.8579                                  & 0.5737                              \\
            & \textit{I} & 0.8684                                  & 0.8737                              & 0.8789                                  & \textbf{0.8579}                     & \textbf{0.8737}                         & \textbf{0.8053}                     & \textbf{0.8632}                         & \textbf{0.6211}                     \\
            \multirow{2}{*}{\textbf{D2}} & \textit{S} & 0.8357                                  & 0.8671                              & \textbf{0.7640}                         & 0.8934                              & \textbf{0.6871}                         & \textbf{0.9161}                     & 0.4423                                  & 0.9493                              \\
            & \textit{I} & 0.8357                                  & 0.8671                              & 0.7413                                  & \textbf{0.8951}                     & 0.5455                                  & 0.8951                              & 0.4423                                  & \textbf{0.9755}                     \\
            \multirow{2}{*}{\textbf{D3}} & \textit{S} & 0.9713                                  & 0.9884                              & \textbf{0.8706}                         & \textbf{0.9917}                     & 0.7675                                  & 0.9801                              & 0.6200                                  & 0.5839                              \\
            & \textit{I} & \textbf{0.9718}                         & \textbf{0.9889}                     & 0.8525                                  & 0.9847                              & \textbf{0.7892}                         & \textbf{0.9810}                     & \textbf{0.7647}                         & \textbf{0.9223}                     \\
            \multirow{2}{*}{\textbf{D4}} & \textit{S} & 0.8723                                  & 0.8532                              & \textbf{0.6465}                         & 0.7544                              & 0.6116                                  & \textbf{0.7281}                     & 0.6221                                  & \textbf{0.7182}                     \\
            & \textit{I} & \textbf{0.8729}                         & 0.8532                              & 0.6425                                  & \textbf{0.7551}                     & \textbf{0.6228}                         & 0.7228                              & \textbf{0.6438}                         & 0.6886                              \\
            \multirow{2}{*}{\textbf{D5}} & \textit{S} & 0.9200                                  & 0.8800                              & 0.9400                                  & 0.5300                              & 0.9400                                  & 0.5200                              & 0.8800                                  & 0.4900                              \\
            & \textit{I} & 0.9200                                  & 0.8800                              & 0.9400                                  & 0.5300                              & 0.9400                                  & \textbf{0.6800}                     & 0.8800                                  & \textbf{0.7100}                     \\
            \multirow{2}{*}{\textbf{D6}} & \textit{S} & 0.9840                                  & 0.9220                              & 0.9820                                  & \textbf{0.7670}                     & 0.9880                                  & 0.6970                              & 0.9880                                  & 0.9810                              \\
            & \textit{I} & 0.9840                                  & \textbf{0.9260}                     & \textbf{0.9850}                         & 0.7330                              & 0.9880                                  & \textbf{0.7370}                     & 0.9880                                  & 0.9810                              \\ \cmidrule(l){2-10} 
        \end{tabular}
    \end{table}

\end{document}